\documentclass[journal,twoside]{IEEEtran}
\usepackage{comment}
\usepackage{amsmath}
\usepackage{amsfonts}
\usepackage{amssymb}
\usepackage{todonotes}
\usepackage{balance}
\usepackage{color}
\usepackage{graphicx}
\usepackage{array}
\usepackage{color,soul}
\usepackage{verbatim}
\usepackage{float}
\usepackage[caption=false]{subfig}

\usepackage[ruled,vlined,linesnumbered]{algorithm2e}
\title{Incremental Deep Neural Network Learning using Classification Confidence Thresholding}
\author{
Justin Leo and Jugal Kalita \\
\textit{Department of Computer Science} \\
\textit{University of Colorado at Colorado Springs} \\
\{jleo, jkalita\}@uccs.edu
}

\begin{document}
    \maketitle
    
    \begin{abstract}
        Most modern neural networks for classification fail to take into account the concept of the unknown. Trained neural networks are usually tested in an unrealistic scenario with only examples from a closed set of known classes. In an attempt to develop a more realistic model, the concept of working in an open set environment has been introduced. This in turn leads to the concept of incremental learning where a model with its own architecture and initial trained set of data can identify unknown classes during the testing phase and autonomously update itself if evidence of a new class is detected. Some problems that arise in incremental learning are inefficient use of resources to retrain the classifier repeatedly and the decrease of classification accuracy as multiple classes are added over time. This process of instantiating new classes is  repeated as many times as necessary, accruing errors.  To address these problems, this paper proposes the Classification Confidence Threshold approach to prime neural networks for incremental learning to keep accuracies high by limiting forgetting. A lean method is also used to reduce resources used in the retraining of the neural network. The proposed method is based on the idea that a network is able to incrementally learn a new class even when exposed  to a \textit{limited number} samples associated with the new class. This method can be applied to most existing neural networks with minimal changes to network architecture.
    \end{abstract}
    
    \section{Introduction}
        When a trained classification model is used, it is realistically deployed in a complex environment that is likely to be dynamic and evolving. In other words, to perform effectively in a realistic environment, a trained network should be able to perform well on classes on which it was trained as well as recognize classes on which it was not trained \cite{draelos2017neurogenesis} \cite{rebuffi2017icarl}. 
        Most  neural classifiers fail to incorporate the possibility of such a complex environment. This may result in a failure of trained neural networks, resulting in inadequate classification. Thus, most neural networks suffer from the problem of inaccurate classification as well as the inability to dynamically learn newly encountered unknown classes.   The objective of this research is to produce a method to adapt the existing neural classification process to allow for incremental class learning while minimizing accumulated error and resource use; this includes limiting network size and training times.
        
        The problem of not being able to handle examples of previously unseen classes can be remedied by endowing a network with the ability to perform open set classification. Open set classification is the ability of a trained classifier to distinguish between known classes and identify potential unknown classes \cite{scheirer2012toward}. This is particularly useful as recognizing unknown classes greatly improves a classifier's accuracy in a complex environment. 
        This research explores incremental learning, the process where a neural classifier continuously learns new classes when encountering  examples from such classes. 
        It repeatedly uses ideas from open set classification to recognize new classes and instantiate them.      
        If a neural network can be primed to expect unknown classes,  it can continuously learn by itself when encountering such unknown data. 
        
        An important factor to consider when developing an incremental learning model is to reduce  the error of classifying samples of currently  known classes mistakenly as belonging to unknown classes. 
        If such errors are made early, they are
         propagated through the learning process, making overall accuracy poor.  Limiting this error, especially early in the process,   allows for continued accurate results \cite{leo2020moving}. In this paper,  we propose a method to modify a few widely used existing classification models and adapt them to work as incremental class learners that are exposed to examples of novel classes in addition to previously learned classes. 
         This  process  depends on being careful in identifying examples as belonging to new classes, in particular in reducing error in mis-recognizing examples of known classes as belonging to unknown classes. 
         When a network potentially identifies the existence of one or more new classes, the proposed method also minimizes the resources used to further train the network. In this paper, we focus on both vision  and natural language classification tasks, and show that our approach builds robust incremental classifiers.
        
        Recent research in deep learning has shown  that  compared to large complex neural networks,  carefully extracted smaller networks are also capable of having similar classification abilities \cite{hassantabar2019scann} \cite{yin2020dreaming}. This is an important factor to consider as the process of incremental learning adds information to a network and naturally makes it bigger. The incremental learning method proposed in this paper focuses on minimizing resource use and thus limits the growth of the network as much as possible.
        
    The main contributions of this paper include  the following: 
    \begin{itemize}
        \item We develop a method to prime neural networks for incremental learning so that they expect to encounter unknown classes.
        \item We present a classification confidence threshold method used to perform open set classification for the purpose of incremental learning, and for the purpose of limiting catastrophic forgetting.
        \item We minimize resource use when incrementally learning new classes for networks.
        \item We develop a new incremental learning metric that evaluates the network at each new incremental learning step.
        
    \end{itemize}
    
     The rest of the paper is organized in the following manner. Section II of the paper discusses related work. Section III introduces our approach to incremental class learning. Section IV presents a new metric for evaluation of incremental class learning. Section V discusses results of several neural network architectures as well as compares the approach with some state-of-the-art approaches. Section VI concludes the paper. 
    
    \section{Related Work}
        The related work is presented in terms of four topics: incremental class learning, open set recognition, node neurogenesis, and optimization of networks.

        Open set incremental learning is a multi-stage process. Leo and Kalita \cite{leo2020moving} build upon the work by Prakhya et al. \cite{prakhya2017open} to propose a technique that can be used by a trained text classifier to recognize unknown classes and to use the identified unknown samples to retrain the trained classifier. This will then include the unknown classes in the network's set of known classes. They modify the softmax layer in a multi-layer convolutional neural network and replace it with an ensemble layer designed to identify novel classes. The ensemble layer is comprised of a voting model between three outlier detection approaches: Mahalanobis Weibull \cite{taguchi2002mahalanobis}, Local Outlier Factor \cite{kriegel2009loop}, and Isolation Forest \cite{liu2008isolation}. The examples of unknown classes identified by the ensemble layer is used to fully retrain the classifier to handle new classes with high accuracy. The approach described in our research uses a similar concept where the softmax layer of a classifier is modified in order to adapt a model for incremental learning.
        
        Incremental learning is a process that incorporates open set classification. Open set classification is the process that identifies the unknown data from the known data. Work on open set classification has been in both NLP \cite{prakhya2017open} \cite{pritsos2013open} as well as Computer Vision \cite{bendale2016towards} \cite{scheirer2012toward}. Open set classification also has an aspect of \textit{confidence}; this is the certainty of the network in determining test data as known or unknown. Dhamija et al. \cite{dhamija2018reducing} propose a method to handle the open set classification problem by developing a new loss function. The new loss function increases the confidence in determining unknown data by maximizing the entropy between unknown and known data samples. The concept of confidence is emphasised in our approach as we develop a novel method to optimize incremental learning.
        
        Neural networks attempt to digitally model the human brain and associated neural functions. Since humans are capable of continuous development and acquisition of novel knowledge, a similar functionality is desirable for a neural network. Based on this concept, Draelos et al. \cite{draelos2017neurogenesis} propose a novel learning method by studying adult neurogenesis by adding new neurons to deep layers of neural networks. These new neurons facilitate the acquisition of previously unseen information and move the model to an incremental learning mode. Using this method, the process of obtaining new knowledge minimizes the utilization of resources. Our approach also utilizes neurogenesis by adding new nodes to grow the incremental learning network's size in order to gradually learn new classes.
   
        A problem often encountered when implementing the incremental learning process is a network's high number of parameters. The fixed architecture, expensive training cost, and density of the model make it difficult to efficiently update the model to accommodate previously unseen classes. The solution to this issue would be to optimize the neural network. Some methods optimize networks in order to reduce resource consumption such as training cost and memory. Dai et al. \cite{dai2019incremental} use a method that prunes unnecessary nodes from networks. Rudd et al. \cite{rudd2017extreme} use a method specific to incremental learning and optimize the specific data samples needed for the multiple incremental learning steps. Resource management and optimization are also part of our work as the approach limits the data used for the retraining steps. Our approach also balances the optimization such that while the training cost is kept minimal, minimal knowledge is lost from the network.

    \section{Approach}
        \begin{figure*}[ht]
            \centering
            \includegraphics[width=\textwidth,height=6cm]{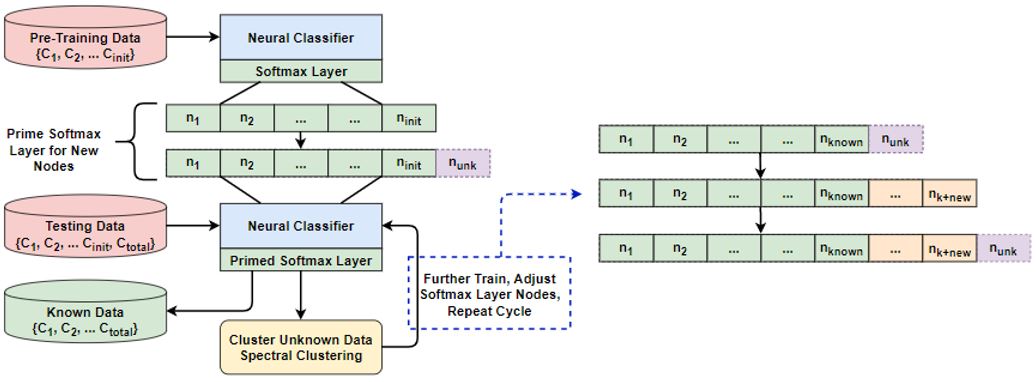}
            \caption{Flowchart model of the Classification Confidence Threshold approach. The classifier's softmax layer is primed to add the potential \textit{unknown data} node if unknown data samples are identified. After the potential unknown data has been identified and clustered, new nodes are added to the softmax layer based on the number of new classes (shown on right of diagram). The classifier is then further trained on the newly labeled data to learn the new classes.}
            \label{fig:approachdgm}
        \end{figure*}
        
        \subsection{Overview}
The proposed approach is called the Classification Confidence Threshold approach as it focuses on optimizing a confidence level for classification. We start with a classification neural network that can  classify the input for a small number of classes. We train it  to achieve high classification accuracy and then the classifier is \textit{deployed} or tested. During deployment, the trained classifier is augmented by adding a  class node at the output layer. This is a \textit{priming} node that will \textit{catch} examples of new classes on which the network was not trained. These \textit{caught} examples are clustered to identify groups unseen by the classifier so far. Output class nodes are instantiated for these groups, augmenting the network, and then discarded. 
Learning of such new classes is repeated till all classes are learned. 

For incremental class learning to work well, there are two conditions that need to be satisfied:
\begin{itemize}
    \item The augmented networks should consistently perform well on examples of trained classes, mapping them to the output nodes corresponding to the trained classes. 
    \item An augmented network should also be such that the extra node catches samples from unknown classes for new class instantiation. 
\end{itemize}
Our training protocol ensures that both of these requirements are  taken care of and the
method can be applied to modify any neural classifier. The approach converts pre-existing classifiers to incremental class learners. 
 
 \subsection{Primary Notations Used}
We first present the primary notations we use in the rest of the paper. Other notations that are used in the paper are introduced when needed. 

We represent a dataset as $\bf D$. 
A dataset $\bf D$ is comprised of a  number of examples, each one of which has a set of features $\bf x$ and a label  $y$. Thus, a dataset ${\bf D} = {<{\bf x}_i, y_i>}, i=1..m$, where $m$ is the total number of examples in the dataset.

Let $\mathcal C$ be the set of classes represented in the dataset $\bf D$. Let there be $n_{total}$ classes. Thus, ${\mathcal C} = \{C_1, \cdots C_{n_{total}} \}$. The dataset can be thought of as the union of a number of sub-datasets, each corresponding to one class in the dataset. In other words,
\begin{equation}
{\bf D} =  {\bf D}_{1} \cup {\bf D}_{2} \cdots   \cup {\bf D}_{n_{total}}
\end{equation}
where ${\bf D}_i$ is the subset of examples that belong to class $C_i$. 
The $m$ examples of the dataset $\bf D$ are usually divided into two disjoint parts, one for training and the other for testing: ${\bf D}^{train}$ and ${\bf D}^{test}$. The training set contains examples of all classes as well as the test set. In particular, we can write
\begin{align}
{\bf D}^{train} &  =  {\bf D}^{train}_{1} \cup{\bf D}^{train}_{2}\cdots  \cdots \cup {\bf D}^{train}_{n_{total}} \\
{\bf D}^{test} &  =  {\bf D}^{test}_{1} \cup{\bf D}^{test}_{2}\cdots  \cdots \cup {\bf D}^{test}_{n_{total}.} 
\end{align}
Here, ${\bf D}^{train}_{i}$ is the subset of training examples from class $C_i$, and ${\bf D}^{test}_{i}$ is the subset of test examples in class $C_i$.

In incremental learning of classes, we first build a classifier $\mathcal K_0$ trained to classify $n_{init}$ number of classes; obviously, we train using training examples of these classes: ${\bf D}^{train}_{1} \cup{\bf D}^{train}_{2}\cdots \cup {\bf D}^{train}_{n_{init}}$. 
Let us assume that the  set of  classes $\mathcal C$ is arranged such that ${\mathcal C} = \{C_1,\cdots C_{n_{init}} \cdots C_{n_{total}} \}$, i.e., the initial classifier $\mathcal K_0$ is built on the first $n_{init}$ classes, and incremental classification continues by learning $n_{incr}$ new classes from the sequence of classes in $\mathcal C$ at a time. $n_{incr}$ is a small positive integer. 
For example, if $n_{incr} =1$, we learn one new class at a time. 
Therefore, we build $\lceil{ 
\frac{n_{total} - n_{init}}
{n_{incr}}} \rceil$ 
classifiers in sequence during incremental learning, given a dataset $\bf D$ with $n_{total}$ classes. We call the sequence of classifiers we build ${\mathcal K}_0 \cdots {\mathcal K}_{\lceil{ 
\frac{n_{total} - n_{init}}
{n_{incr}}} \rceil}$.

For this paper, we also need to define some terms frequently used. A \textit{known class} is a class previously seen by the classifier and the associated data are \textit{known data}. An \textit{unknown class} is a class not previously seen by the classifier and the associated data are \textit{unknown Data}. Lastly, \textit{noise data} are a random mix of data from known classes.

\subsection{The Algorithm in Brief}
We build an initial classifier that is trained on the initial $n_{init}$ classes. 
This initial classifier is then tested on examples of the classes on which it was trained as well as training examples of the next $n_{incr}$ classes, which we consider unknown at this time.  We use an extra output node we inject into the network, to \textit{catch} the examples of the unknown classes. These caught examples of the $n_{incr}$ unknown classes are clustered into $n_{incr}$ clusters, and these clusters with the caught examples are used as synonyms of the  unknown classes. The network is updated by adding $n_{incr}$ output nodes corresponding to these new classes. Thus, the classifier evolves via a number of iterations:  one iteration consists of training the classifier network on examples of a  certain number of classes, catching examples of unknown classes, clustering these examples, and using the clusters to instantiate new classes. It is quite likely that the evolving classifier makes mistakes in catching or recognizing examples of unknown classes in such iterations, or in classifying examples of known classes as unknown. Any error made in an early iteration propagates through the later iterations, decreasing the overall accuracy of the classifier. Thus, care must be taken to reduce the amount of error committed during the iterations when new classes are learned progressively, especially in earlier iterations. 

As discussed in the previous paragraph, we discover that it is not necessary to train the evolving classifier on all examples of the  previously unknown classes; a few correctly chosen examples obtained from the catching and clustering process are enough. The process of catching and clustering makes the classifier aware of the presence of examples of previously unknown classes. To fortify the classifier so that its performance does not degrade, the evolving classifier is retrained on the caught unknown data samples as well as some noise data; this helps reduce \textit{catastrophic forgetting}. 
 
 As we can see, in the first iteration of the incremental class learning process, the classifier is trained on all training examples of the first $n_{init}$ classes. However, when it learns the next $n_{incr}$ classes, it is not trained on all examples of these new classes; only those examples that are caught by the current classifier with its limited knowledge of classes. 

 The protocol for incremental learning that we follow  is given precisely in Algorithm 1, with an accompanying  illustration in Figure \ref{fig:approachdgm}. Lines 4-6 initialize the algorithm by setting  up the first $n_{init}$ classes as currently known classes. It also sets the rest of the $n_{total} - n_{init}$ classes as unknown classes to be learned incrementally. The initial classifier trained on the first $n_{init}$ classes is called $\mathcal{K}_0$. The {\em for} loop in lines 7-17 builds a sequence of classifiers $\mathcal{K}_1, \cdots$ to incrementally learn all $n_{total}$ classes. The steps inside the loop go through one such incremental class learning process and are discussed in subsequent subsections. 
 
 As an example, if for a classification task, $n_{total}=50, n_{init}=5$ and $n_{incr}=1$, the algorithm builds an initial classifier $\mathcal{K}_0$ to classify the  first 5 classes $C_1, \cdots, C_5$. It then incrementally augments this initial classifier by building a  sequence of classifiers $\mathcal{K}_1, \cdots, \mathcal{K}_{45}$, classifying $6,\cdots,50$ classes, respectively. Classfier $\mathcal{K}_1$ classifies classes $\{ C_1, \cdots, C_6\}$ whereas classifier $\mathcal{K}_{45}$ classifies all classes in $\mathcal{C}=\{C_1,\cdots,C_{50}\}$. 

 Our method can be applied to modify any neural classifier, including high-preforming pre-existing ones, and can move the existing classification models to an incremental learning state without much decrease of accuracy.

\begin{algorithm*}
        \SetAlgoLined
        \textbf{Input:} A dataset $\bf D$ with examples from a set of classes $\mathcal{C} = \{ C_1, \cdots C_{n_{init}}, \cdots, C_{n_{total}} \}$
        
        \textbf{Input:} A small positive integer $n_{incr}$, which is the number of new classes learned in each iteration

        \textbf{Output:} Incrementally Trained Classifier $\triangle \mathcal{K}$ that recognizes all $n_{total}$ classes
        
        $\mathcal{C}^{known}_0 \leftarrow \{ C_1, \cdots C_{n_{init}} \}$,      $\mathcal{C}^{unknown}_0 \leftarrow \mathcal{C} - \mathcal{C}^{known}_0$ initial sets of known and unknown classes

        $\triangle \mathcal{K}_0 \leftarrow$ Train classifier on examples of $\mathcal{C}^{known}_0$ classes  from $\bf D$: 
                {\color{black} Training data = $ {\bf D}^{train}_{1} \cup{\bf D}^{train}_{2}\cdots \cup {\bf D}^{train}_{n_{init}} $}

        \For{$l=1$ to $\lceil{ \frac{n_{total} - n_{init}}{n_{incr}}} \rceil$ } 
            {
            $\triangle \mathcal{K}^{primed}_l \leftarrow$
                prime classifier network $\triangle \mathcal{K}_{l-1}$ with an unknown output class node $\circledcirc C_l^{unk}$
                                            
            $\mathcal{C}^{newunknown}_l \leftarrow 
             \{C_{n_{init}+(l-1)n_{incr}} \cdots C_{n_{init}+l\;n_{incr}} \}$, new unknown classes to test with
            
            $\mathcal{C}^{alltested}_l \leftarrow \mathcal{C}^{known}_{l-1} \cup  \mathcal{C}^{newunknown}_l$

            Test $\triangle \mathcal{K}^{primed}_l$ on examples for classes belonging to  $\mathcal{C}^{alltested}$:
            {\color{black} Testing Data = $\{ \bigcup_{C_i \in \mathcal{C}^{known}_{l-1}} {\bf D}^{test}_{i} \} 
            \cup \{ \bigcup_{C_i \in \mathcal{C}^{newunknown}_l} {\bf D}^{train}_i \}$; test data for known classes, all data for new unknown}  
            
             {\color{black} ${\bf D}^{knownsampled}  = \{{\bf D}^\prime_1  \cdots {\bf D}^\prime_{n_{init}+(l-1)n_{incr} -1 }  \} \leftarrow$ 
           sample a small number of examples from the training subsets of the known classes to obviate catastrophic forgetting}
                                   
           {\color{black} ${\bf D}^{newunknown} = \{{\bf D}^\prime_{n_{init}+(l-1)n_{incr}} \cdots {\bf D}^\prime_{n_{init}+l\;n_{incr}} \} \leftarrow$ Recognize examples of unknown classes $\mathcal{C}^{newunknown}_l$ using Algorithm 2, which takes as input all examples associated with priming node $\circledcirc C
            ^{unk}_l$ and obtains $n_{incr}$ clusters }

            $\triangle \mathcal{K}_l \leftarrow \triangle \mathcal{K}^{primed}_l - \circledcirc C_l^{unk} +$ output nodes $\circledcirc OutNodes$ for $\mathcal{C}^{newunknown}_l$, newly augmented network
           
           $\mathcal{C}^{known}_l \leftarrow \mathcal{C}^{known}_{l-1} \cup \mathcal{C}^{newunknown}_l $
           
            $\mathcal{C}^{unknown}_l \leftarrow \mathcal{C}^{unknown}_{l-1} - \mathcal{C}^{newunknown}_l$
            
            $\triangle \mathcal{K}_l \leftarrow$ Train  $\triangle \mathcal{K}_l$ on $\mathcal{C}^{known}_l$ with {\color{black} data from ${\bf D}^{newunknown} \cup {\bf D}^{knownsampled}$}
        }

        \textbf{return} $\triangle  \mathcal{K}_l$
            \caption{Overview of the Classification Confidence Threshold Approach: We use $\circledcirc$ to indicate network nodes, and $\triangle$ to indicate networks for additional clarity.}

    \end{algorithm*}

        \subsection{The Idea of Priming to Learn New Classes}
        In this paper, we present a method for  incremental class learning  that repeatedly performs open set classification as it increases the number of classes it knows. While there have been other approaches to incremental learning, such approaches typically involve fully retraining the classifier or extensive layer editing, making them  resource expensive as the network may have a multitude of layers and a large training step \cite{dai2019incremental,leo2020moving,castro2018end,li-etal-2019-incremental}.
        To reduce the amount of resource usage during incremental learning as well as to improve efficiency and accuracy of incremental learning, we  use the concept of priming the network's output layer to  discover and learn new classes during testing or deployment.  Our hypothesis is that a classification network that is primed for incremental class learning  minimizes  re-training steps. 
        
        The  term \textit{Priming} in Psychology  refers to  ``a technique whereby exposure to one stimulus influences a response to a subsequent stimulus, without conscious guidance or intention" \cite{bargh2000studying}. 
        
        In other words, without conscious effort, certain tasks or actions make other tasks or actions easier to perform in the near future. When we talk about priming here, we refer to \textit{positive priming}. Positive priming \cite{reisberg2007cognition} 
        means that the first stimulus or action activates parts of a particular representation or association in memory just before carrying out an action or task. The representation is already partially activated when the second stimulus  is encountered or the second action is  performed,  and as a result less additional work is necessary. Motivated by this idea of priming in psychology, we intend to implement it in  our neural network for incremental class learning. We hypothesize that as a neural network learns to classify a certain number of classes, it is already primed to learn a small number of additional classes. In other words, it already has  some of the neural structure needed to learn a small number of related classes. 
    
        Priming takes place in each iteration when the currently trained network is tested with $n_{incr}$ new classes, with a view to recognizing unknown classes for incorporation.

        \subsection{Implementing Priming in the Final Layer}
       Most neural networks use a softmax layer as shown in Equation \ref{eqn:softmax} for classification; in the scenario where a softmax layer is not normally used, a softmax layer can be added to the end of the network. The softmax layer takes as input a sequence of numbers, produces a corresponding sequence of numeric outputs  such that each output is  between 0 and 1, all the numbers add up to 1, and the larger input numbers become bigger so that the big ones stand out still bigger. 

    The first stage of the incremental learning approach is to modify the softmax layer of the classifier.  The concept of open set classification implies that there is always the possibility of encountering examples of novel classes in any testing or deployment environment. 
    Due to this reason, the softmax layer is modified  by adding an extra node such that there is always the possibility of encountering $n_{incr}$ potential unknown classes. To prime the network, we add  one extra node after the initial training process; this extra node is also initialized by slightly balancing the weights of the other softmax nodes. The balancing causes a minimal change of the weights as to not drastically interfere with the trained knowledge of the network. The purpose of the priming is to facilitate the confidence threshold approach as detailed in Algorithm 2. 
    The  softmax layer's output is shown in Equation \ref{eqn:softmax}.
\begin{equation}
    softmax (\vec{x}, i) = \frac{e^{x_i}} {\sum_{i=1}^k e^{x_j }}
    \label{eqn:softmax}
\end{equation}
We simply add an extra priming node to the softmax layer. This happens in Line 8 of Algorithm 1.
     
     \subsection{Determining Weights to Prime Node}
        After the network is initially trained, it is primed for potential unknown classes. The weights to the new unknown node need to be initialized before the model can utilize them for testing and further training. Ideally, these nodes should not disrupt the information already contained in the network; to accomplish this, a small portion of each weight from the existing softmax nodes is allocated to the weights of the new nodes as shown in Equation \ref{eqn:new_node_weight}. If new classes are identified, these new nodes' weights are updated in the further training step. Here \textit{k} is the number of pre-trained nodes and \textit{N} is an arbitrary large number, for our experimentation we use  \textit{N} = 1000 as that produces the most consistent results. 
        \begin{equation}
            weight(n_{k+1}:n_c) = \frac{\sum_{i=1}^k n_i} {N}
            \label{eqn:new_node_weight}
        \end{equation}
        

       If data belonging to one or more novel classes are found, these new nodes become the classification nodes for the data and a new node \textit{n\textsubscript{new}} is added.
       
\subsection{Training the Network}
        After the softmax layer is primed for adding unknown nodes, the second step is the testing process with a combination of known trained data and unknown data. Normally this step would involve performing open set classification and obtaining the unknown data samples and retrain the classifier with these samples. 
   
\subsubsection{Issues in Training for Incremental Class Learning}  
        A serious incremental learning problem is the \textit{erroneous classification of known data examples as unknown}; \cite{leo2020moving};
        incremental learning is a continuous process and this means the error propagates over time as the incremental cycle continues. To remedy this problem, we propose using the idea of a \textit{confidence threshold}. As shown in Algorithm 2, the approach only obtains unknown data if there is a significant low classification confidence in the softmax layer. The confidence threshold utilizes the sample's softmax probabilities to determine if the sample should be classified as unknown. This process only obtains a fraction of the total unknown data samples, but in our experimentation we find the neural network is capable of finding the new unknown class with a limited number of samples. This means the approach utilizes a limited instantiation of open set classification in order to improve incremental learning performance. The confidence threshold can be tuned to maximize results based on the architecture. 
        \begin{algorithm*}
            \SetAlgoLined
            \textbf{Input:} ${\bf D}^{test}$, Testing data from Line 11 of Algorithm 1; data consists of all test data for the known classes, and all data (test and train) for new unknown classes 
            
           \textbf{Input:} CT Value $c$ from Equation \ref{eqn:Conf_Thresh}; $n_{incr}$: number of unknown classes; $n_{known}$: number of current known classes
          
             \textbf{Output:} ${\bf D}^{newunknown}$, a set of examples recognized as belonging to new unknown classes
         
            ${\bf D}^{newunknown} \leftarrow \phi$, empty set
            
            $UnknownSamples \leftarrow \phi$, empty set

            \For{$t \in {\bf D}^{test}$ ($t$ is a test sample)}
            {
                ${\bf p}^t = \{p^t_1, \cdots, p^t_{n_{known}+1} \} \leftarrow $ Compute softmax probabilities that $t \in C_i, i =1, \cdots n_{known}+1$; $n_{known}$ is the number of currently known classes, the last node is the priming node
                
                ${\bf p}^\prime \leftarrow$ Set of probabilities $\bf p$ without $max (\bf p)$, the highest value
            
                \eIf{$max(\bf p) > (avg(\bf p^\prime) * c)$}
                {${flag}^t  \leftarrow 0;$ $t$ is in a known class}
                {${flag}^t = 1;$ $t$ is in an unknown class\\
                $UnknownSamples \leftarrow  UnknownSamples \cup \{ t \}$}

            }
            
            ${\bf D}^{newunknown} \leftarrow $ cluster $UnknownSamples$ into $n_{incr}$ clusters

            \textbf{return} ${\bf D}^{newunknown}$
                \caption{Unknown Sample Detection with Confidence Threshold}
        \end{algorithm*}
        
        The Classification Confidence Threshold approach, as proposed in this paper is a concept that has been briefly explored before in open set classification for computer vision. The approach, described in Equation \ref{eqn:EVM_Unknown} \cite{rudd2017extreme}, utilizes a confidence value to identify open set data.

        \begin{equation}
            y^*=\begin{cases}
                argmax_{l\in\{1,...,M\}} P(C_l\mid x^{'}) &
                \text{if } P(C_l\mid x^{'}) \geq
                \sigma\\
                \text{``unknown"} & \text{otherwise}
            \end{cases}
            \label{eqn:EVM_Unknown}
        \end{equation}
        
        The problem with this method is that it is designed more for open set classification rather than incremental learning. As mentioned before, our approach is designed to obtain high accuracy for incremental learning rather than open set classification results. Rudd et al. \cite{rudd2017extreme} also show an approach for incremental learning; however, an additional step is added to limit the resources required for multiple reclassification steps. With the approach described in this paper, the additional step is not required as the Classification Confidence Threshold algorithm already limits the unknown data size for retraining and reclassification. Thus, an efficient low resource utilization method is developed for incremental learning.
        
\subsubsection{Clustering, Handling Multiple Unknown Classes}   
        One additional problem that arises is the potential for multiple unknown classes to arise during the testing process. To address this problem, we use a clustering approach to determine if data from multiple classes is present. The clustering algorithm used is Spectral Clustering \cite{stella2003multiclass}. Spectral clustering was shown to perform well with open set identification \cite{leo2020moving}. If the clustering process produces more than one unknown class, the softmax layer is modified to add more unknown nodes corresponding to the new classes; since the network is already primed to add additional nodes, this step is relatively quick and does not cause much performance degradation.

\subsubsection{Retraining the Enhanced Network}
     \begin{figure}[ht]
        \centering
        \includegraphics[width=9cm,height=5.5cm]{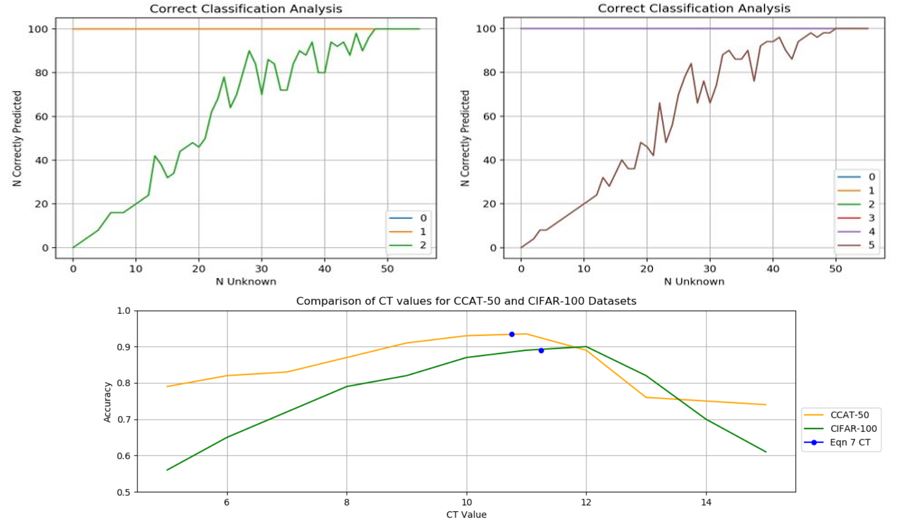}
        \caption{
        Confidence Threshold analysis using Multi-Layer CNN and data classification. These plots show accuracies of re-training an already pre-trained network; this test was conducted on all the datasets resulting a similar pattern. 
        Both plots have a certain number of classes fully pre-trained and a new class is incrementally added; the first plot shows the addition of class 2 and the second plot shows the addition of class 5. The x-axis shows the number of samples used in the re-training process, since some classes are already pre-trained, the accuracies for these stay at 100\%. The new class being added has the testing accuracy increase as more of its associated samples are used for the re-training. Two observations are identified by these plots: for a given dataset approximately the same number of samples are needed to fully learn a new class (this number changes based on dataset), and a network is able to recognize/learn a new class with a fewer number of samples. The third plot shows how changing the confidence threshold affects the results and Equation \ref{eqn:Conf_Thresh} 
        's output for the addition of one class.}
        \label{fig:confidence_plots}
    \end{figure}
        The final stage is further training the classifier with the identified data samples of unknown classes so the new classes can be learned. Using our low classification confidence threshold approach, the network is further trained with the samples of the unknown classes. In a realistic scenario, the network would be exposed to both unknown data as well as known data, so random mixed data samples (referred to as noise data samples) are added during the retraining phase. The noise data also help the network to not forget information about the old data in the further training step. After further training, the classifier is able to find most of the new classes' data in the testing stage even if the training only included a limited number of the associated classes' samples. 
        One feature of the proposed approach is the low resource consumption in order to achieve incremental learning. Even though an additional training step is needed in the process, the new training data is greatly limited through the proposed approach, and so this additional time is minimal in comparison to other approaches such as fully retraining the classifier or iterative node addition through the network's deep levels. Through this process, the model moves towards an incremental learning structure while maintaining high classification accuracy results.

     \subsection{Confidence Threshold Determination}
            The main idea that the Confidence Threshold approach addresses is that the primary source of error for incremental learning is erroneous labeling of new training data. So to address this problem, a confidence threshold needs to be calculated such that the unknown data selected for training is truly unknown as well as being labeled correctly. Figure \ref{fig:confidence_plots} shows that we have more variability when determining the confidence threshold value because the number of samples needed to learn a new class is approximately the same regardless of iteration (based on dataset), and that a network is able to learn a new class even if all the new samples are not used for training. 
            
            Some other aspects of incremental learning models also have an effect on the confidence threshold determination. The first aspect is that more error is introduced in each iteration if multiple classes are added in a single iteration. The second aspect is that all incremental learning models slowly loose accuracy over multiple iterations. The third aspect is that incremental learning accuracy is dependent on the regular classification accuracy of the selected dataset. Based on the main idea of the Confidence Threshold approach and the incremental learning aspects, Equation \ref{eqn:Conf_Thresh}
            was formed to calculate thresholds. $n_{iner}$ is the number of classes added per iteration, $l$ is the iteration number, $A_{init}$ is the base classification accuracy of the entire dataset, $a$ is an initial constant value that's variable (use either 9 or 10). The bias is only used in rare cases where the network needs a greater confidence threshold, only applies to the Transformer and BERT network. The confidence threshold value shows the largest softmax weight has to be greater than the threshold times the average of the other weights. Figure \ref{fig:confidence_plots}
            's third plot shows a comparison of different confidence threshold values for the first iteration, Equation \ref{eqn:Conf_Thresh} 
            produces approximately the best value.
        \begin{equation} 
            CT = \bigg[\frac{(n_{iner} * l)+a}{A_{init}}\bigg]
            \label{eqn:Conf_Thresh}
        \end{equation}
    \section {Evaluation Methods}
        Normal metrics used for classification evaluation are accuracy and f1-score; however, we need a method for evaluating incremental learning classification. Incremental learning is a process where a classification network learns new classes over iterations of learning, so each iteration must be evaluated for accuracy and combined to a total score. For the purposes of this paper, we define one iteration as one cycle of discovering unknown class samples and fine training on the newly labeled data. The reasoning behind evaluating the classifier between each iteration is the propagated error through each iteration \cite{leo2020moving}. If the incremental learning classifier is only evaluated at the end of all the iterations, the final accuracy score might be misleading because the final accuracy score may be high, but one of the previous iteration's accuracy scores may have been low. This would mean the classifier has not fully learned one or more of the incrementally added classes. We propose a metric called \textit{Incremental Learning Accuracy} or \textit{ILA}, as shown in Equation \ref{eqn:ILA}, that takes into account each iteration's accuracy and produces a final accuracy for the classifier. In Equation \ref{eqn:ILA}, $m^i_{correct}$ is the total number of correctly classified test samples and $m^i_{total}$ is the total number of test samples in iteration \textit{i}. The score is a type of accuracy, the values range from 0 to 1.
        
        \begin{equation}
        ILA = \frac{1}{l}
            \sum_{i=1}^{l}  
                \bigg[\frac{m^i_{correct}}{m^i_{total}} 
                \bigg], l=Iteration Count
            \label{eqn:ILA}
        \end{equation}
        
        Other equations and approaches have also been proposed for open set classification that involve assessing if the classifier can find all unknown class samples \cite{leo2020moving}, \cite{rudd2017extreme}, and \cite{bendale2016towards}. 
        The incremental learning approach closest to this paper's approach uses Incremental Class Accuracy (ICA) \cite{leo2020moving}. This metric assesses how well the clustering step needed for open set classification is performed by averaging homogeneity of clusters, completeness of clusters, and full encapsulation of the open set data. 
        While this metric is useful for open set classification using clustering, this method is designed such that it is not necessary for all the unknown class samples to be found, but rather our goal is to limit the propagated error through the learning process. Thus, the proposed ILA scores show the results of this work better. 
        Along with the ILA scores, the work also shows the accuracy change over time through the iterations of the incremental learning process.
        
    \section {Experiments and Results}
        \begin{table*}[ht]
          \centering
          \renewcommand{\arraystretch}{1.2}
          \begin{tabular}{|c|c||c|c|c|c|c|c|}
            \hline
            Model & Classes Added & CCAT-50 & Amazon & CIFAR-100 & EMNIST & ImageNet & Caltech-101\\
            \hline
            Multi-Layer CNN \cite{leo2020moving} & 1 & 0.935 $\pm$0.021 & 0.937 $\pm$0.026& 0.894 $\pm$0.035& 0.967 $\pm$0.008& 0.818 $\pm$0.032& 0.861 $\pm$0.024\\
            
                             & 2 & 0.879 $\pm$0.023& 0.911 $\pm$0.030& 0.842 $\pm$0.045& 0.934 $\pm$0.009& 0.739 $\pm$0.038& 0.739 $\pm$0.023\\
                             
                             & 3 & 0.830 $\pm$0.038& 0.827 $\pm$0.036& 0.813 $\pm$0.045& 0.863 $\pm$0.009& 0.711 $\pm$0.042& 0.710 $\pm$0.028\\
                             
                             & 5 & 0.781 $\pm$0.049& n/a & 0.787 $\pm$0.051& n/a & 0.654 $\pm$0.045& 0.668 $\pm$0.044\\
                             
                             & 10 & 0.628 $\pm$0.051& n/a & 0.710 $\pm$0.049& n/a & 0.554 $\pm$0.047& 0.576 $\pm$0.045\\
                             
            \hline
            LeNet-5 & 1 & 0.930 $\pm$0.022& 0.912 $\pm$0.025& 0.913 $\pm$0.037& 0.962 $\pm$0.005& 0.803 $\pm$0.029& 0.760 $\pm$0.024\\
                    & 2 & 0.879 $\pm$0.025& 0.882 $\pm$0.035& 0.854 $\pm$0.042& 0.922 $\pm$0.010& 0.717 $\pm$0.035& 0.729 $\pm$0.021\\
                    & 3 & 0.818 $\pm$0.035& 0.831 $\pm$0.037& 0.816 $\pm$0.041& 0.893 $\pm$0.011& 0.707 $\pm$0.041& 0.707 $\pm$0.023\\
                    & 5 & 0.774 $\pm$0.047& n/a & 0.751 $\pm$0.048& n/a & 0.618 $\pm$0.040& 0.667 $\pm$0.042\\
                    & 10 & 0.618 $\pm$0.055& n/a & 0.706 $\pm$0.050& n/a & 0.562 $\pm$0.043& 0.560 $\pm$0.049\\
            \hline
            ResNet-18 & 1 & 0.871 $\pm$0.028& 0.891 $\pm$0.028& 0.887 $\pm$0.038& 0.970 $\pm$0.007& 0.821 $\pm$0.035& 0.799 $\pm$0.028\\
                      & 2 & 0.816 $\pm$0.028& 0.827 $\pm$0.034& 0.815 $\pm$0.045& 0.936 $\pm$0.013 & 0.722 $\pm$0.037& 0.726 $\pm$0.028\\
                      & 3 & 0.781 $\pm$0.033& 0.798 $\pm$0.037& 0.789 $\pm$0.045& 0.887 $\pm$0.014 & 0.720 $\pm$0.044& 0.723 $\pm$0.029\\
                      & 5 & 0.728 $\pm$0.048& n/a & 0.745 $\pm$0.044& n/a & 0.650 $\pm$0.041& 0.669 $\pm$0.045\\
                      & 10 & 0.615 $\pm$0.057& n/a & 0.704 $\pm$0.048& n/a & 0.576 $\pm$0.042& 0.551 $\pm$0.046\\
             \hline
            Transformer & 1 & 0.907 $\pm$0.015& 0.909 $\pm$0.018& n/a & n/a & n/a & n/a\\
                        & 2 & 0.856 $\pm$0.022& 0.855 $\pm$0.023& n/a & n/a & n/a & n/a\\
                        & 3 & 0.823 $\pm$0.042& 0.808 $\pm$0.042& n/a & n/a & n/a & n/a\\
                        & 5 & 0.800 $\pm$0.045& n/a & n/a & n/a & n/a & n/a\\
                        & 10 & 0.628 $\pm$0.054& n/a & n/a & n/a & n/a & n/a\\
            \hline
            BERT & 1 & 0.929 $\pm$0.014& 0.926 $\pm$0.015& n/a & n/a & n/a & n/a\\
                 & 2 & 0.884 $\pm$0.024& 0.871 $\pm$0.024& n/a & n/a & n/a & n/a\\
                 & 3 & 0.861 $\pm$0.039& 0.808 $\pm$0.040& n/a & n/a & n/a & n/a\\
                 & 5 & 0.803 $\pm$0.046& n/a & n/a & n/a & n/a & n/a\\
                 & 10 & 0.628 $\pm$0.048& n/a & n/a & n/a & n/a & n/a\\
            \hline
            \hline
            Test & Stat Output & CCAT-50 & Amazon & CIFAR-100 & EMNIST & ImageNet & Caltech-101\\
            \hline
            ANOVA Test & F-Statistic & 0.201& 0.442& 0.131& 0.676& 0.043& 0.105\\
             & P-Value & 0.935& 0.776& 0.878& 0.543& 0.957& 0.901\\
            \hline
          \end{tabular}
          \caption{ILA mean and standard deviation for each dataset and tested models based on Equation 4. Scores are based on the results shown in Figures }\ref{fig:Results_Text} and \ref{fig:Results_Image}. These scores show the accuracy of the models after all iterations of incremental learning when adding different number of classes. Only the large datasets could be tested with 5 and 10 classes. The ANOVA test was also calculated for each dataset comparing the results of each model type; the low F-Statistics and high P-Values show the groups have similar means, and this shows the Confidence Threshold method performs consistent for each model.
        \end{table*}
        \begin{table}[ht]
          \centering
          \renewcommand{\arraystretch}{1.2}
          \begin{tabular}{|c||c|c|c|}
            \hline
            Approach & CIFAR-100 & ImageNet & Caltech-101\\
            \hline
            FT & 0.215 & 0.205 & 0.422\\
            \hline
            AL & 0.299 & 0.316 & 0.478\\
            \hline
            EWC & 0.337 & 0.268 & 0.475\\
            \hline
            MAS & 0.342 & 0.293 & 0.459\\
            \hline
            iCaRL & 0.472 & 0.475 & 0.589\\
            \hline
            FN & 0.665 & 0.513 & 0.639\\
            \hline
            \textbf{CCT} & \textbf{0.682} & \textbf{0.535} & \textbf{0.661}\\
            \hline
          \end{tabular}
          \caption{ILA mean scores comparing proposed and existing approaches. Initially train 50 classes with increments of 5 classes per iteration. The CCT approach is the proposed method.}
        \end{table}
        
    \subsection{Datasets Used} 
        Since this paper explores classification on both text and image data, two datasets are used for text and four datasets are used for images. The following datasts are used for text: CCAT-50 \cite{houvardas2006n}, Amazon Reviews Data \cite{ni2019justifying}. The following datasets are used for images: CIFAR-100 \cite{Krizhevsky09learningmultiple}, EMNIST \cite{cohen2017emnist}, ImageNet-Subset \cite{deng2009imagenet}, Caltech-101 \cite{fei2004learning}.
        
    \subsection{Models Tested}
         For the Classification Confidence Threshold algorithm, specific deep neural network architectures are tested. Since the approach is being tested for both text and image data, some models chosen are targeted to work with both forms of data. Most of the models being tested are forms of the CNN architecture as they have proven to be state-of-the-art for open set image classification \cite{bendale2016towards}. These CNN models are also tested with the text data as they also work well for open set text classification \cite{higashinaka2014towards}. For text CNN models, the word2vec \cite{mikolov2013distributed} model is used to obtain word vector embeddings. 
         Another architecture type commonly used for text data is a Transformer model \cite{vaswani2017attention}. Most state-of-the-art natural language processing models are based on Transformer models such as BERT \cite{Devlin2019BERTPO}; so the proposed incremental learning algorithm is applied and tested on the base Transformer model and the BERT model.

         The models being tested are: a multi-layer CNN model \cite{ciresan2011flexible} as described in \cite{leo2020moving} and \cite{prakhya2017open}, a LeNet-5 model \cite{lecun1998gradient}, a ResNet-18 model \cite{he2016deep}, a Transformer model \cite{vaswani2017attention} for textual data, and a BERT model \cite{Devlin2019BERTPO} for textual data.
        \begin{figure*}[ht]
            \centering
            \includegraphics[width=\textwidth,height=4.8cm]{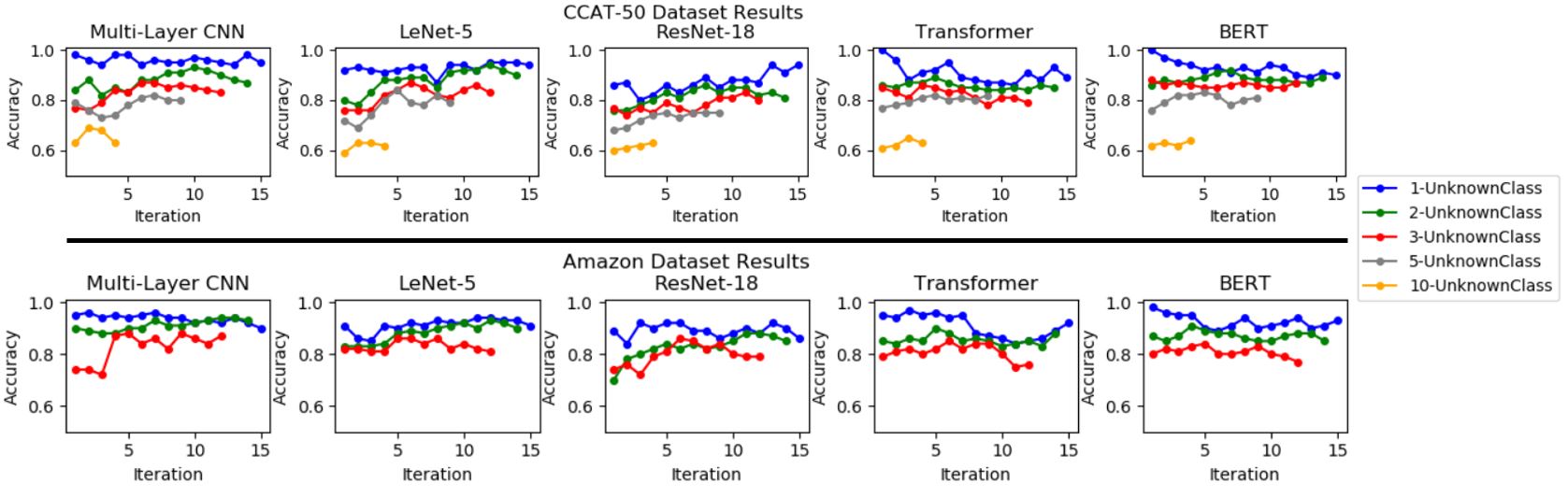}
            \caption{Incremental learning results for the tested models for the Text datasets. The plots show the classification accuracy over each iteration of the incremental learning cycle. Each model is tested by adding x unknown classes per iteration.}
            \label{fig:Results_Text}
        \end{figure*}
        \begin{figure*}[ht]
            \centering
            \includegraphics[width=\textwidth,height=4.8cm]{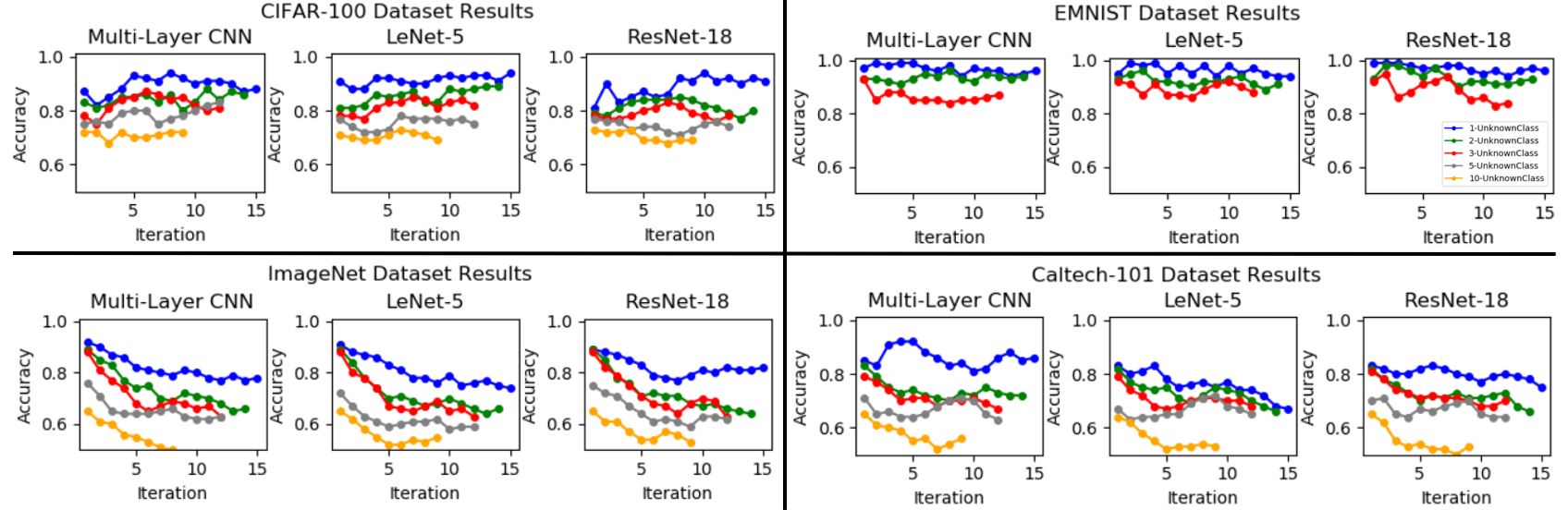}
            \caption{Incremental learning results for the tested models for the Image datasets. The plots show the classification accuracy over each iteration of the incremental learning cycle. Each model is tested by adding x unknown classes per iteration.}
            \label{fig:Results_Image}
        \end{figure*}
         \begin{figure*}[ht]
            \centering
            \includegraphics[width=\textwidth,height=3.7cm]{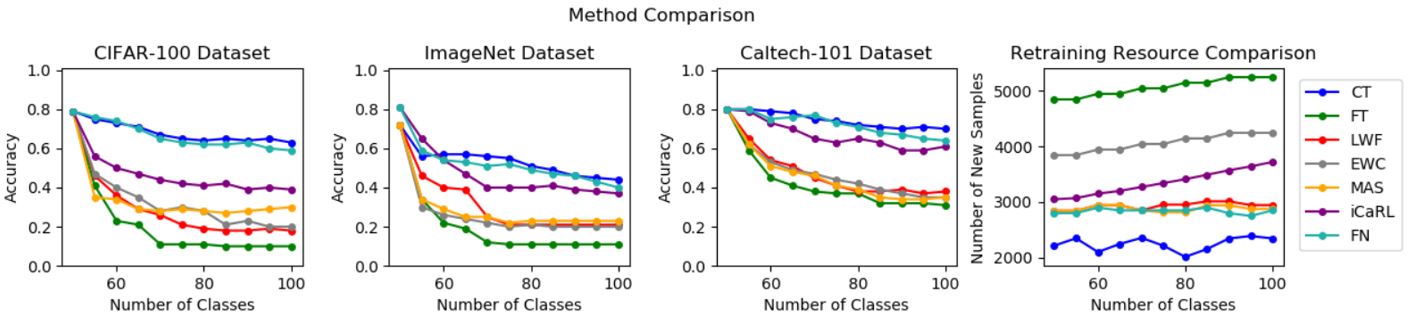}
            \caption{Comparison to results of existing approaches. Each network was initially trained on 50 classes and 5 classes are added per iteration. The last plot shows average number of data samples processed in the neurogenesis step for each approach at each iteration.}
            \label{fig:Results_Method_Comparison}
        \end{figure*}
        
    \subsection{Comparison to Existing Approaches}
        This paper focuses on incremental learning and specifically addresses the problem of erroneous misclassification of unknown data as known data. A common problem identified in all incremental learning approaches is a neural network's tendency to forget data as more knowledge is added over time through all incremental learning processes. While forgetting knowledge is often the most observable problem in incremental learning, it is often greatly amplified due to erroneous misclassification problem. This is because when a new data sample gets classified as known data, it impairs the network to correctly classify data for the known class in future learning and testing iterations. This accidental error during classification of data is easily construed as forgetting as the true class is now misrepresented by the network. The Classification Confidence Threshold approach described in this paper greatly helps reduce the network's forgetting behavior as it is specifically designed to reduce erroneous misclassification. To test this, we compare the proposed approach to other incremental class learning techniques specifically designed to reduce forgetting.
    
        \noindent \textbf{Finetuning (FT)} \cite{hoffer2015deep}: This method uses networks trained using the triplet loss function and are used to compute distance comparisons between classes to determine if a data sample is known.
        
        \noindent \textbf{Alignment Loss (AL)} \cite{li2017learning}: This method matches softmax outputs of previously trained models to current data to determine if samples belong to known classes.
        
        \noindent \textbf{Elastic Weight Consolidation (EWC)} \cite{kirkpatrick2017overcoming}: This method focuses on keeping a network's parameters similar when adding a new class to the network.
        
        \noindent \textbf{Memory Aware Synapses (MAS)} \cite{aljundi2018memory}: This method calculates an importance score for each parameter of a network based on how sensitive the predicted output can change the parameter values.
        
        \noindent \textbf{Incremental Classifier Representation Learning (iCaRL)} \cite{rebuffi2017icarl}: This method is designed to learn over time by learning strong classifiers and data representations simultaneously.
        
        \noindent \textbf{FearNet (FN)} \cite{kemker2018fearnet}: This method is designed to incrementally learn by using a dual-memory design where knowledge is consolidated from a network to replicate a brain inspired long-term memory storage.
    
    \subsection{Incremental Class Learning Results}
        Incremental learning is a process of continuous training and classification through iterations. After the initial training step, where the classifier learns an initial $n_{init}$ number of classes, each iteration tests on new $n_{incr}$ unknown classes along with the previously learned classes. The testing phase separates out the predicted examples of unknown data and is retrained only on the new classes mixed with some noise data of previously known classes. As discussed earlier, a neural network is able to classify samples of new classes even if only trained on a limited number of associated samples; so the goal of the process is to limit the error produced by classifying samples of known classes mistakenly as unknown classes. 

        To test if this process proves successful for incremental learning, a neural network model must first be trained on a known set of classes and incrementally tested as novel classes are added. The first test trains on $n_{init}=5$ classes and adds $n_{incr}=1, 2, 3, 5, 10$ classes incrementally over a maximum of 15 iterations to calculate the incremental learning accuracy per iteration. Results are shown in Table 1 and Figures \ref{fig:Results_Text}, \ref{fig:Results_Image}. Note only the large datasets can be tested for adding 5 and 10 classes each iteration. 
        Table 1 also performs the ANOVA statistical test to show statistical significance between mean values. The ANOVA test was performed for each of the six datasets comparing the results of each  of the five model types. The low F-Statistics and high P-Values in each column show the groups have similar means, and this shows the Confidence Threshold method performs consistently for each model.
        
        The next test compares our proposed Classification Confidence Threshold (CCT) approach with other approaches as described in the previous section. For these tests, the focus is mainly on limiting the network forgetting as new classes are added. The networks are trained initially on 50 classes and 5 classes are added incrementally; results are shown in Figure \ref{fig:Results_Method_Comparison} and Table 2. The incremental learning tests performed all use the ResNet-18 architecture. The last plot in Figure \ref{fig:Results_Method_Comparison} shows the average number of new data samples used for the retrain/neurogenesis step where new classes are learned. The proposed approach is designed to only select a reduced number of samples, so the re-train step each iteration is more efficient than that of the compared methods as the algorithm processes a vastly reduced number of data samples.
        
        The accuracy values do not change much for each iteration and this shows that our method is working, and the propagated error is greatly limited per iteration. However, the approach has the greatest error in the clustering process; so when more classes are added at once, the error in the clustering step increases and this is observed in the results. Apart from these incremental accuracy graphs, we also calculate our proposed Incremental Learning Accuracy (ILA) scores for each model and the results are shown in Table 1. High ILA scores show us that the networks were successfully able to incrementally learn the existence of new classes and classify the associated samples with high accuracy. Our proposed approach also performs better than the other approaches and reduces forgetting. 
        
    \section{Conclusion}
    This paper develops the Classification Confidence Threshold method to adapt existing neural networks to move to incremental learning. Our experiments show success with NLP and CV classification tasks. The approach modifies a network and primes the softmax layer for neurogenesis. The new classes are added to the network with minimal use of resources while maintaining a high classification accuracy. The retraining step only includes the few data samples identified by the thresholding approach; so this step along with the node addition step are quicker than the initial training. Incremental learning requires multiple iterations to form a continuous cycle; so to evaluate the iterations we also developed a new metric to analyze the accuracies within each iteration.
    
    The major issue identified and addressed from prior approaches is the propagated error through each cycle. To limit the propagated error, it is important that only the samples of unknown classes are identified correctly for retraining. We also find that a network trained on a few samples only can be used to identify all samples of the corresponding class; so using the Classification Confidence Threshold approach we find only a limited number of correct samples so a high classification accuracy is maintained. The main limitation is within the clustering process for multiple unknown class identification. Thus, the highest accuracy results are obtained when only adding a single class per iteration. Using these methods we demonstrate an efficient incremental learning process.

    \balance
    \bibliographystyle{IEEEtran}
    \bibliography{bibliography}
\end{document}